\def\year{2020}\relax
\documentclass[letterpaper]{article} 
\usepackage{aaai20}  
\usepackage{times}  
\usepackage{helvet} 
\usepackage{courier}  
\usepackage[hyphens]{url}  
\usepackage{graphicx} 
\usepackage{graphicx}  
\def\levelvariable{setup}
\def\Levelvariable{Setup}
\def\levelvariables{setups}

\title{Fine-Grained Argument Unit Recognition and Classification}
\author{
    Dietrich Trautmann\textsuperscript{\textdagger}, 
    Johannes Daxenberger\textsuperscript{\textdaggerdbl}, 
    Christian Stab\textsuperscript{\textdaggerdbl}, 
    Hinrich Sch{\"u}tze\textsuperscript{\textdagger}, 
    Iryna Gurevych\textsuperscript{\textdaggerdbl}\\
    \textsuperscript{\textdagger}Center for Information and Language Processing (CIS), 
    LMU Munich, Germany\\
    \textsuperscript{\textdaggerdbl}Ubiquitous Knowledge Processing Lab (UKP-TUDA), 
    TU Darmstadt, Germany\\
    {\tt dietrich@trautmann.me; inquiries@cislmu.org}\\
    \url{http://www.ukp.tu-darmstadt.de}\\
}

\long\def\eat#1{}

\def\tabref#1{Table~\ref{tab:#1}}
\def\tablabel#1{\label{tab:#1}\label{p:#1}}
\def\secref#1{Section \ref{sec:#1}}
\def\seclabel#1{\label{sec:#1}}
\def\eqref#1{Eq.~\ref{eqn:#1}}

\newcommand\newcite[1]{\citeauthor{#1} (\citeyear{#1})}

\usepackage{booktabs}
\usepackage{multirow}
\usepackage{hhline}

\newcounter{notecounter}

\newcommand{\enoteson}{\long\gdef\enote##1##2{{
                \stepcounter{notecounter}
                            {\large\bf
                                   \hspace{1cm}\arabic{notecounter} $<<<$ ##1: ##2
                                   $>>>$\hspace{1cm}}}}}
\enoteson

\begin{document}


\maketitle

\begin{abstract}
    Prior work has commonly defined argument retrieval from heterogeneous document collections as a sentence-level classification task.
    Consequently, argument retrieval suffers both from low recall and from sentence segmentation errors making it difficult for humans and machines to consume the arguments.
    In this work, we argue that the task should be performed on a more fine-grained level of sequence labeling.
    For this, we define the task as Argument Unit Recognition and Classification (AURC).
    We present a dataset of arguments from heterogeneous sources annotated as \emph{spans of tokens} within a sentence, as well as with a corresponding stance.
    We show that and how such difficult argument annotations can be effectively collected through crowdsourcing with high inter-annotator agreement.
    The new benchmark, AURC-8, contains up to 15\% more arguments per topic as compared to annotations on the sentence level.
    We identify a number of methods targeted at AURC sequence labeling, achieving close to human performance on known domains.
	Further analysis also reveals that, contrary to previous approaches, our methods are more robust against sentence segmentation errors.
    We publicly release our code and the AURC-8 dataset.\footnote{\url{https://github.com/trtm/AURC}}
\end{abstract}


\section{Introduction}
\label{sec:intro}
Argumentation and reasoning are fundamental human skills.
They play a major role in education, daily conversations, as
well as in many professional contexts including journalism,
politics and the law.
Argumentative skills are trained, for example, in the context of (public) debates, which are an essential part of democratic societies.
Argument mining (AM), i.e., the
processing of argumentative structures in natural language
using automatic methods \cite{Peldszus2013}, has recently gained considerable attention
in the AI community \cite{ijcai2018-766,Lippi:2016:AMS:3016100.3016319,DBLP:conf/aaai/NguyenL18}. 
AM requires sophisticated reasoning about controversial subject matters well beyond mere syntactic or semantic understanding.
In recent research on AM, two radically different paradigms have evolved:
closed-domain \textit{discourse-level} AM seeks to identify the argumentative structure of a debate or an argumentative text (e.g., a student essay).
In contrast, \textit{information-seeking} AM aims to identify self-contained argumentative statements relevant to the given topic from any source.
The goal of this kind of AM is to identify a broad and diverse set of arguments, ideally reflecting different viewpoints and aspects of the topic of interest.
Most approaches to automatic discourse-level AM \cite{Wyner2010,Stab2014a} rely on the \emph{claim-premise} model, which refers to the basic units of an argument as a controversial proposition (claim) and connected evidence (premise).
The claim-premise model is the basis for the analysis of complex argumentative texts, which can be solved by AI-based automatic methods with success \cite{kuribayashi-etal-2019-empirical}.

\begin{figure}
	\centering
	\includegraphics[width=0.3\textwidth]{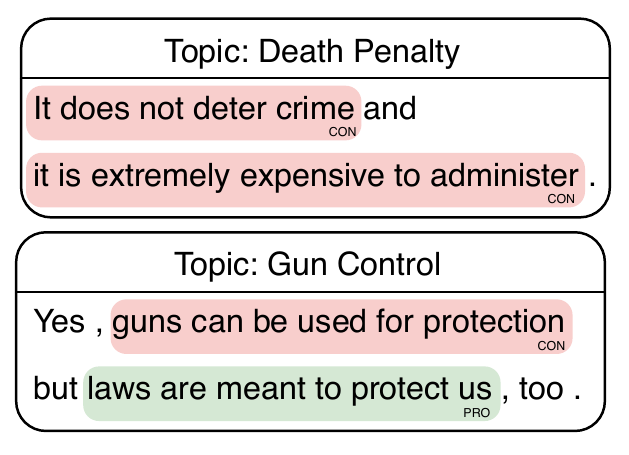}
	\caption{Annotation of argumentative spans and
	stance. Each of the two sentences contains two arguments.}
	\label{fig:example}
\end{figure}

\begin{table*}\centering
\scriptsize
\scalebox{1.1}{
	\begin{tabular}{llrrrrrrrrr}
    \toprule
        \# & topic                  &
        \#sentences & \#candidates & \#final & \#arg-sent & \#arg-unit & increase in \% & \#non-arg \\
    \midrule
        T1 & abortion               &
        39,083 & 3,282 & 1,000 & 424 & 458 & +8.02 & 576 \\
        T2 & cloning                &
        30,504 & 2,594 & 1,000 & 353 & 380 & +7.65 & 647 \\
        T3 & marijuana legalization &
        45,644 & 6,351 & 1,000 & 630 & 689 & +9.37 & 370 \\
        T4 & minimum wage           &
        43,128 & 8,290 & 1,000 & 630 & 703 & +11.59 & 370 \\
        T5 & nuclear energy         &
        43,576 & 5,056 & 1,000 & 623 & 684 & +9.79 & 377 \\
        T6 & death penalty          &
        32,253 & 6,079 & 1,000 & 598 & 651 & +8.86 & 402 \\
        T7 & gun control            &
        38,443 & 4,576 & 1,000 & 529 & 587 & +10.96 & 471 \\
        T8 & school uniforms        &
        40,937 & 3,526 & 1,000 & 713 & 821 & +15.15 & 287 \\
        \midrule
        &  total 		    	&
 314,568 & 39,754 & 8,000 & 4,500 & 4,973 & +10.51 & 3,500  \\
    \bottomrule
    \end{tabular}
}
\caption{Sentences in the selection process and final corpus size.
\#arg-sent: argumentative sentences.
\#arg-unit: argumentative units.
increase in \%: increase of \#arg-unit compared to \#arg-sent.}
\label{tab:corpus_stats}
\end{table*}

The claim-premise model is, however, hardly applicable to
text that does not contain an explicit argumentative structure (i.e., the majority of textual data) and, thus, has not been applied with much success to heterogeneous document collections \cite{habernal2017}.
Information-seeking AM, which allows to detect arguments in sources that are not inherently argumentative, has been suggested as a remedy to this problem.
It solves the following task:
given a controversial claim (``topic''), detect supporting or opposing statements from (potentially) relevant sources.
In this context, an argument is usually defined as a  short
text -- or \emph{span} --  providing evidence or reasoning about the topic, supporting or opposing it \cite{Stab2018b}.
This can be seen as a flat version of the claim-premise model, where the topic is an implicit claim and the argument is a premise that supports or attacks the claim. 
Previous studies on information-seeking AM working with heterogeneous document collections have restricted arguments to be sentences.
This assumption has been partly justified by the difficulty of ``unitizing'', i.e., of segmenting a sentence into meaningful units for argumentation tasks \cite{Stab2018b}.
A few studies also allowed free text fragments as arguments \cite{aharoni-etal-2014-benchmark}, however, their arguments are extracted from very restricted domains (Wikipedia articles) with clear relevance to the topic.
Consequently, it remains unclear whether such argument spans can be extracted from large and heterogeneous document collections with sufficient accuracy. 

Information-seeking AM as defined above has thus not been applied to both highly heterogeneous
document collections \textit{and} on the token level.
By identifying subsentence token spans as arguments for given topics in a large web crawl, we show that this task is feasible.
Furthermore, addressing the unitizing problem, we show how the required training data can be created in a scalable manner and with high agreement using crowdsourcing. 
We call this task \emph{Argument Unit Recognition and
Classification (AURC)}.\footnote{A reviewer pointed out that
neither argument unit recognition nor argument unit
classification are novel subtasks and so the need for introducing a
new abbreviation for their combination may be
questioned. However, we are interested in the specific
scenario of information-seeking AM for highly heterogeneous
collections and for
arguments segmented on the token level (i.e., fine-grained segmentation). This (in our view)
novel scenario justifies a new abbreviation and a new name.}
Labeling arguments on the token level has several advantages.
First and foremost, it prevents merging otherwise separate
arguments into a single argument (e.g., for the
topic \textit{death penalty} in Figure \ref{fig:example}),
enabling us to retrieve
a larger number of arguments  (up to 15\% more in our data).
Second, it can handle two-sided argumentation in a single sentence adequately (e.g., for the topic \textit{gun control} in Figure \ref{fig:example}).
It is also more robust against errors in sentence segmentation.
Finally, downstream applications like fact checking that need to reason over a set of evidences require decomposing complex sentences into simpler argument units.
\newcite{DouglasWalton2017} argue that any kind of argument mining in which parts of a given text are re-used to form new arguments is closely related to argument invention \cite{Levy2014}. 
This view points towards further applications of our approach in the field of rhetoric.


We make the following contributions in this paper.  First,
we propose a slot-filling approach to argument annotation
and show that it effectively collects token-level
annotations through crowdsourcing with high inter-annotator
agreement. Based on this approach, we construct AURC-8 by applying a sampling
strategy to a large web crawl.  Second, we present a number
of methods for solving the AURC task and evaluate them on the AURC benchmark.  Finally, we show that
our token-level model does not depend on correct sentence
boundary identification.


\section{Related Work}

Previous work on AM in AI and Natural Language Processing divides into
discourse-level AM \cite{Palau2009,Stab2014a}
and information-seeking AM \cite{Shnarch2018,Wachsmuth2017,Hua2017,Stab2018b}, as explained in Section~\ref{sec:intro}.
Our work is in line with the latter: we model arguments as self-contained pieces of information that can be verified as relevant arguments for a given topic with no or minimal surrounding context.
As opposed to previous work on information-seeking AM that extracted token-level arguments from Wikipedia articles \cite{Levy2014}, we apply our annotation schema and experiments to a much broader and noisier collection of web documents.

\newcite{Ajjour2017} compare argumentative unit segmentation
approaches on the token level across three corpora.
They use a feature-based approach and various architectures for segmentation and find that BiLSTMs work best on average.
In contrast to our work, they study argumentation from the discourse-level perspective, i.e., they do not consider topic-dependency and do not account for argument stance.
\newcite{peldszus-stede-2015-joint} use elementary discourse units as starting point to classify argumentative discourse units (ADUs).
As opposed to their work, our approach does not rely on discourse parsing to identify ADUs. 
Furthermore, \newcite{peldszus-stede-2015-joint} do not consider topic-dependency of arguments. 

Multiple previous studies have shown that annotating arguments from the discourse-level perspective is very challenging for heterogeneous document collections. 
\newcite{habernal2017} used Toulmin's schema of argumentation on several text genres including web data, resulting in rather low inter-annotator agreement scores (Krippendorff's $\alpha_u$ = 0.48).
\newcite{miller2019} apply the claim-premise model to customer reviews with similarly low agreement ($\alpha_u$ roughly between 0.4 and 0.5).
Using the information-seeking
perspective, \newcite{Stab2018b} prove that arguments can be
annotated with sufficient agreement in heterogeneous sources
on the sentence level (Fleiss’ $\kappa$=0.72).
In our work, we propose a slot filling approach to transfer their findings to token-level arguments.
Previously, \newcite{Reisert2018} achieved good agreement determining complex reasoning structures 
with a set of highly specialized slots referred to as \textit{argument templates}.
Compared to their work, our slot filling approach is simpler, but it is applicable to heterogeneous sources and topics in a crowdsourcing environment.


\section{Corpus Creation}
Collecting annotations on the token level is challenging.
First,  the unit of annotation needs to be clearly defined.
This is  straightforward for tasks with short spans (sequences of words) such as named entities, but much harder for longer spans -- as in the case of argument units.
Second, labels from multiple annotators need to be merged into a single gold standard.\footnote{One could also learn from ``soft'' labels, i.e., a distribution created from the votes of multiple annotators. However, this does not solve the problem that some annotators deliver low quality work and their votes should be outscored by a (hopefully) higher-quality majority of annotators.}
This is also more difficult for longer sequences because simple majority voting over individual words will likely create invalid (e.g., discontinuous or grammatically incorrect) spans.
In the following, we propose solutions to these problems and describe the selection of sources, sampling and annotation for our novel argument unit dataset, AURC-8.

\subsection{Data Source}
We used the February 2016 Common Crawl archive,\footnote{\url{http://commoncrawl.org/2016/02/february-2016-crawl-archive-now-available/}} which was indexed with Elasticsearch\footnote{\url{https://www.elastic.co/products/elasticsearch}} following \newcite{stab2018argumentext}.
Since we want to know, whether knowledge transfer to unknown topics is possible with few annotated topics, we limited the selection to \newcite{Stab2018b}'s eight topics (cf.~Table~\ref{tab:corpus_stats}).
This also increases comparability with related work.
The topics are general enough to have good coverage in Common Crawl.
They are also controversial and hence a good choice for argument mining with an expected diverse set of supporting and opposing arguments.

\subsection{Retrieval Pipeline}
For document retrieval, we queried the indexed data for \newcite{Stab2018b}'s topics and collected the first $500$ results per topic ordered by their document score (\emph{doc\_score}) from Elasticsearch; a higher \emph{doc\_score} indicates higher relevance for the topic.
For each document, we retrieved the corresponding WARC file at the Common Crawl Index.\footnote{\url{http://index.commoncrawl.org/CC-MAIN-2016-07}}
From there, we downloaded and parsed the original HTML document for the next steps of our pipeline; this ensures reproducibility.
Following this, we used  justext\footnote{\url{http://corpus.tools/wiki/Justext}} to remove  HTML boilerplate.
We used spacy\footnote{\url{https://spacy.io/}} to segment the resulting document into  sentences and sentences into tokens.
We only consider sentences with the number of tokens in the range $[3,45]$.

\subsection{Sentence Sampling}
\seclabel{sampling}

We pre-classified the selected sentences with a sentence-level argument mining model following \newcite{stab2018argumentext} and available via the ArgumenText Classify API.\footnote{\url{https://api.argumentsearch.com}}
The API returns for each sentence (i) an argument confidence
score \emph{arg\_score} in $[0.0,1.0)$ (we discard sentences
with \emph{arg\_score}$< 0.5$), (ii) the stance on the
sentence level ({PRO} or {CON}) and (iii) the stance confidence score \emph{stance\_score}.
This information was used together with the \emph{doc\_score} to rank sentences for a selection in the following crowd annotation process.
See Table~\ref{tab:corpus_stats} for a detailed overview of the number of extracted sentences, as well as the number of candidates for the following ranking approach.
We convert scores for documents, arguments and stance   into ranks $d_{i}$, $a_{i}$ and $s_{i}$.
We then sum the three ranks to get an aggregated score for each sentence: $\mbox{agg}_{i} = d_{i} + a_{i} + s_{i}$.
Sentences are divided by topic and pre-classified stance and ranked according to $\mbox{agg}_{i}$, for each combination separately.
We then go down the ranked list selecting each sentence with probability $0.5$ until the target size of $n=500$ per stance and topic is reached; otherwise we do  additional passes through the list. We adopted this probabilistic selection to ensure diversity -- otherwise a long document with high relevance score at the top of the list might dominate the dataset with its sentences.
Table \ref{tab:corpus_stats} gives the final dataset creation statistics.

\subsection{Crowd Annotations}
\seclabel{crowdannotations}
Our  goal  is to develop a scalable approach to annotate
argument units on the token level.
Given that arguments need to be annotated with regard to a
specific topic, training data for different topics needs to
be created.  As has been shown by previous work on
information-seeking AM \cite{Shnarch2018,Stab2018b}, crowdsourcing (on
the sentence level) can be used to obtain reliable annotations for argument mining datasets.
However, as outlined above, token-level annotation significantly increases the difficulty of the annotation task.
We distinguish between PRO (supporting) and CON (opposing) arguments and we use NON for non-argumentative text spans.
Thus, we have a sequence labeling task with three classes: PRO, CON and NON.
Our main question was: can we achieve sufficiently high
agreement among untrained crowd workers for this task?

We use  the $\alpha_u$ \textbf{agreement measure} \cite{Krippendorff2016a} in this work.
It is designed for annotation tasks that involve unitizing textual continua -- i.e., segmenting continuous text into meaningful sub-units -- and measuring chance-corrected agreement in those tasks.
It is also a good fit for argument spans within a sentence: typically these spans are long and the context is a single sentence that may contain any type of argument and any number of arguments.
\newcite{Krippendorff2016a} define a family of $\alpha$-reliability coefficients that improve upon several weaknesses of previous $\alpha$ measures.
From these, we chose the $\alpha_{u_{nom}}$ coefficient, which also takes into account agreement on ``blanks'' (non-arguments in our case) -- since agreement on non-argument spans (including sentences without any arguments) is important as well and should not be ignored by the measure.

To determine agreement, we initially carried out an \textbf{in-house expert study} with three graduate employees (who were trained on the task beforehand) and randomly sampled 160 sentences (10 per topic and stance) from the overall data.
In the first round, we did not impose any restrictions on the span of words to be selected, other than that the selected spans should be the shortest self-contained spans that form an argument.
This resulted in unsatisfactory agreement ($\alpha_{u_{nom}}$ = 0.51, average over topics), one reason being inconsistency in selecting argument spans (median length of arguments ranged from nine to 16 words among the three experts).

In a second round, we therefore decided to restrict the spans that could be selected by applying a slot filling approach enforcing valid argument spans that match a template.
We use the template: ``\emph{$<$TOPIC$>$} should be supported/opposed, because \emph{$<$argument~span$>$}''.
The guidelines specify that the resulting sentence must be
grammatically correct.\footnote{Strict adherence to
grammatical correctness would require that all spans are clauses.
A reviewer asked if
there are examples of non-clauses in our data.
Although these cases are rare in our gold standard, they
do occur. Examples 
include
noun phrases like ``risky animal experiments''
(against the topic ``cloning'')
and
clauses that
are missing the subject, which is usually to be equated with
the topic, e.g., 
``have no real impact on improving student achievement''
(for the topic ``school
uniforms'').
These
argument units are easily comprehended by
humans even though they are not complete clauses in the strict
grammatical sense.}
Although this new setup increased the length
of spans and reduced the total number of arguments selected,
it increased consistency of spans substantially (min and max median lengths were now 15 and 17).
Furthermore, the agreement between the three experts rose to $\alpha_{u_{nom}}$ = 0.61 (average over topics).
Compared to other studies on token-level argument mining \cite{EckleKohler2015,Li2017,Stab2014a}, this score is in an acceptable range and we deem it sufficient to proceed with crowdsourcing.
The averaged macro-$F_1$ over all pairs of expert annotations is 0.76 (referred to as human performance in \tabref{freq_base}).

In our \textbf{crowdsourcing setup}, workers could select one or multiple spans, where each span's permissible length is between one token and the entire sentence.
Workers had to select at least one argument span and its stance (supporting/opposing); alternatively, if they could not find an argument span, they had to solve a simple math problem.
We employed two quality control measures: a qualification test and periodic attention checks.\footnote{Workers had to be located in the US, CA, AU, NZ or UK, with an acceptance rate of 95\% or higher. Payment was \$0.42 per HIT, corresponding to US federal minimum wage (\$7.25/hour).}
On an initial batch of 160 sentences, we collected votes from nine workers.
To determine the optimal number of workers for the final
study, we did majority voting on the token level (ties  broken as non-arguments) for both the expert study and workers from the initial crowd study.
We artificially reduced the number of workers (1-9) and calculated percentage overlap averaged across all worker combinations (for worker numbers $\leq$ 9).
Whereas the overlap was highest with $80.2\%$ at nine votes, it only dropped to $79.5\%$ for five votes (and decreased more significantly for fewer votes).
We deemed five votes to be an acceptable tradeoff between quality and cost.
The agreement between experts and crowd in the five-worker setup is $\alpha_{u_{nom}}$ = 0.71, which is substantial \cite{Landis1977}.

The final \textbf{gold standard} labels on the 8000 sampled sentences were determined using a variant of Bayesian Classifier Combination \cite{Kim2012IBCC}, referred to as IBCC in \newcite{Simpson2018}'s  modular framework for Bayesian aggregation of sequence labels.
This method has been shown to yield results superior to majority voting or MACE \cite{hovy2013learning}.
After manual inspection of the resulting gold standard, we merged all consecutive segments of the same stance, to form a gold standard with coherent segments separated by at least one other token label (most of the time NON).

\subsection{Dataset Splits}
\label{sec:splits}
We create two different dataset splits.
(i) An in-domain split.
This lets us evaluate how models perform on known vocabulary and data distributions.
(ii) A cross-domain split.
This lets us evaluate how well a model generalizes for unseen topics and distributions different from the training set.\footnote{We use \textit{cross-domain} rather than \textit{cross-topic} \cite{Stab2018b} here, as the former is the more common term.}
In the cross-domain setup, we defined topics T1-T5 to be in the train set, topic T6 in the dev set and topics T7 and T8 in the test set.
For the in-domain setup, we excluded topics T7 and T8 (cross-domain test set), and used the first $70\%$ of the topics T1-T6 for train, the next $10\%$ for dev and the remaining $20\%$ for test.
The samples from the in-domain test set were also excluded in the cross-domain train and dev sets.
As a result, there are 4000 samples in train, 800 in dev and 2000 in test for the cross-domain split; and 4200 samples in train, 600 in dev and 1200 in test for the in-domain split.
Our definition of the two splits guarantees that train/dev sets (in-domain or cross-domain) do not overlap with test sets.
The assignment of sentences to the two splits is released as part of AURC-8.

\subsection{Dataset Statistics}
The resulting dataset, AURC-8, consists of 8000 annotated sentences; 3500 (43.8\%) of which are non-argumentative.
The 4500 argumentative sentences are divided into 1951
(43.4\%) single PRO argument sentences, 1799 (40.0\%) single
CON argument sentences and 750 (16.7\%) sentences that are
many possible combinations of PRO and CON arguments with up
to five single argument units in a sentence.
The total number of argumentative segments is 4973. 
Thus, due to the token-level annotation
the number of arguments is 
higher by 10.5\% compared to what it would be with a
sentence-level approach (4500).
If we propagate the label of a sentence to all its tokens,
then 100\% of tokens of an argumentative sentence are argumentative.
This ratio drops to 69.9\% in our token-level setup, reducing the amount of non-argumentative tokens otherwise incorrectly labeled as argumentative in a sentence.


\section{Experimental Setup}
\seclabel{expset}

We train and evaluate in two different setups: token-level
and sentence-level.
In the \emph{token-level \levelvariable}, models are
trained and evaluated on the gold standard as is. 
For the \emph{sentence-level \levelvariable}, we use
a  sentence-level gold standard that is modified by converting
token-level gold standard or predictions  to a prediction for the sentence as follows.
If only NON occurs, the sentence is labeled NON. If NON and
only PRO (resp.\ only CON) 
occurs, PRO (resp.\ CON) is chosen.
If both PRO and CON occur, the more frequent label of the two is assigned.
In the few exceptional cases where PRO and CON are equally frequent, one of them is chosen by chance.

\subsection{Evaluation Measures}
\seclabel{evalmeas}

We compute three different evaluation measures of AURC:
\textbf{token $F_1$}, \textbf{segment $F_1$}  
and \textbf{sentence $F_1$}.

\textbf{Token $F_1$} is defined as the average of the three class $F_1$'s for 
PRO, CON and NON, each computed for all tokens in the
evaluation set. This is the core evaluation measure of our
work since our goal is argument retrieval
at a more fine-grained level than prior work.\footnote{To illustrate this point, we use the term ``argument retrieval'' rather than ``information-seeking AM'' in the following.}

To compute \textbf{segment $F_1$} for a sentence, 
we look at all pairs $<$gold segment $g$,  predicted segment $p$$>$ and 
compute: \[r= \frac{|g \cap p|}{\max(|g|,|p|)}\]
If  $r > 0.5$, then we count $p$ as a true positive if the labels are the same.
We do this only for PRO and CON segments, so the correct recognition of NON segments 
(which is of less benefit in an argument retrieval scenario) is not rewarded.
However, if the sentence contains no PRO or CON span and no PRO or CON span is predicted, 
then we set $F_1$ for this sentence to 1.0.
Thus, an AURC model is rewarded for recognizing a sentence that is entirely NON.
Macro-averaged segment $F_1$ is then the average of segment $F_1$'s for all 
sentences in the evaluation set.

Segment $F_1$ is  a measure of evaluating fine-grained
argument mining that is complementary to token $F_1$. It is
more forgiving than token $F_1$ since segment boundaries
do not have to be recognized exactly. At the same time, it
punishes cases where tokens are mostly recognized correctly,
but the segments that a downstream application may need
(e.g., a user interface displaying arguments) are not. For example,
a gold segment may be split in half by a single NON
token, resulting in two incomplete predicted segments.

For \textbf{sentence} $F_1$, a single label is predicted for each sentence.
The three class $F_1$'s for PRO, CON and NON are  computed for the sentences 
in the evaluation set and averaged to yield the overall sentence $F_1$.
Gold standard labels for sentences are produced as described above (\emph{sentence-level \levelvariable}).

\subsection{Methods}
\seclabel{methods}
We test and extend models that have recently well performed on AM and sequence labeling \cite{ReimersSBDSG19}.

\textbf{Baseline.}
The majority baseline labels each token of the input sentence with NON, the most frequent class.

\textbf{FLAIR.}
We use FLAIR \cite{akbik2018contextual}, a recent neural sequence labeling 
library with state-of-the-art performance on many sequence labeling tasks.
For the token-level \levelvariable, we use the FLAIR SequenceTagger
with a CRF and with
stacked character embeddings \cite{lample2016neural}.
For the sentence-level \levelvariable, we use the FLAIR TextClassifier.
In both cases, the selected model is a
BiLSTM \cite{hochreiter1997long}  and we use GloVe word embeddings \cite{pennington2014glove} and FlairEmbeddings 
(provided by the library).

\textbf{BERT.}
\label{sec:bert}
We include two  pretrained models of
BERT\footnote{\url{https://github.com/huggingface/pytorch-pretrained-bert}} \cite{devlin2018bert}
as a recent state-of-the-art model that achieves impressive
results on many tasks including sequence labeling:
BERT\textsubscript{BASE} (cased, base) and
BERT\textsubscript{LARGE}
(cased, large,
with whole-word-masking).
A third model adds a
CRF \cite{lafferty2001conditional} as a task specific
sequence labeling layer:
BERT\textsubscript{LARGE}+CRF.
We finetune BERT's parameters
(including the CRF) for the AURC task.
Since we work in the information-seeking paradigm of AM,
the topic (i.e., the search query) is of key importance.
A span of words that is a supporting argument (PRO) for ``nuclear energy'' will 
generally be NON for the topic ``gun control''.
Thus we provide BERT  with an additional input, the topic, separated 
by a [SEP] token from the input sequence.
We found that this additional topic information was 
consistently helpful for AURC.
Token- and sentence-level \levelvariables\ for the BERT models only differ with regard to gold standard data.


\def\bigtablesep{-0.05cm}

\begin{table*}[!h]\centering
\scriptsize
\scalebox{1.05}{
\begin{tabular}{|l||c||c||c|}
    \hhline{|-||-||-||-|}
    \begin{tabular}{lrl}
         &  & \multicolumn{1}{r}{} \\
        \midrule
         &  & \multicolumn{1}{r}{} \\
         &  & \multicolumn{1}{r}{} \\
        \midrule
         &  & \multicolumn{1}{l}{model} \\
        \midrule
        \multirow{5}{*}{\rotatebox[origin=c]{90}{in-domain}}
         & 4 & majority baseline \\[\bigtablesep]
         & 5 & FLAIR \\[\bigtablesep]
         & 6 & BERT\textsubscript{BASE} \\[\bigtablesep]
         & 7 & BERT\textsubscript{LARGE} \\[\bigtablesep]
         & 8 & BERT\textsubscript{LARGE}+CRF \\
        \midrule
        \multirow{5}{*}{\rotatebox[origin=c]{90}{cross-domain}}
         & 9 & majority baseline \\[\bigtablesep]
         &10 & FLAIR \\[\bigtablesep]
         &11 & BERT\textsubscript{BASE} \\[\bigtablesep]
         &12 & BERT\textsubscript{LARGE} \\[\bigtablesep]
         &13 & BERT\textsubscript{LARGE}+CRF \\
        \midrule
         &14 & human performance \\
    \end{tabular}&
    \begin{tabular}{cc|cc}
        \multicolumn{4}{c}{token $F_1$} \\
        \midrule
        \multicolumn{2}{c|}{token-level} & \multicolumn{2}{c}{sentence-level} \\[\bigtablesep]
        \multicolumn{2}{c|}{\levelvariable} & \multicolumn{2}{c}{\levelvariable} \\
        \midrule
        dev & test & dev & test \\
        \midrule
        .258 & .254 & .258 & .254 \\[\bigtablesep]
        .642 & .613 & .442 & .473 \\[\bigtablesep]
        .702 & .654 & .591 & .581 \\[\bigtablesep]
        .732 & .683 & \textbf{.671} & \textbf{.627} \\[\bigtablesep]
        \textbf{.743} & \textbf{.696} & .637 & .622 \\
        \midrule
        .245 & .240 & .245 & .240 \\[\bigtablesep]
        .419 & .433 & .388 & .401 \\[\bigtablesep]
        .554 & .563 & .431 & .474 \\[\bigtablesep]
        .604 & .596 & \textbf{.550} & \textbf{.544} \\[\bigtablesep]
        \textbf{.615} & \textbf{.620} & .505 & .519 \\
        \midrule
        \multicolumn{4}{c}{.763} \\
    \end{tabular}&
    \begin{tabular}{cc|cc}
        \multicolumn{4}{c}{segment $F_1$} \\
        \midrule
        \multicolumn{2}{c|}{token-level} & \multicolumn{2}{c}{sentence-level} \\[\bigtablesep]
        \multicolumn{2}{c|}{\levelvariable} & \multicolumn{2}{c}{\levelvariable} \\
        \midrule
        dev & test & dev & test \\
        \midrule
        .478 & .463 & .478 & .463 \\[\bigtablesep]
        .661 & .623 & .473 & .492 \\[\bigtablesep]
        .666 & .628 & \textbf{.615} & \textbf{.601} \\[\bigtablesep]
        .749 & .709 & .599 & .567 \\[\bigtablesep]
        \textbf{.750} & \textbf{.724} & .552 & .547 \\
        \midrule
        .394 & .379 & .394 & .379 \\[\bigtablesep]
        .457 & .458 & .402 & .401 \\[\bigtablesep]
        .504 & .508 & .445 & \textbf{.473} \\[\bigtablesep]
        .653 & .626 & \textbf{.487} & \textbf{.473} \\[\bigtablesep]
        \textbf{.681} & \textbf{.649} & .456 & .464 \\
        \midrule
        \multicolumn{4}{c}{.709} \\
    \end{tabular}&
    \begin{tabular}{cc|cc}
        \multicolumn{4}{c}{sentence $F_1$} \\
        \midrule
        \multicolumn{2}{c|}{token-level} & \multicolumn{2}{c}{sentence-level} \\[\bigtablesep]
        \multicolumn{2}{c|}{\levelvariable} & \multicolumn{2}{c}{\levelvariable} \\
        \midrule
        dev & test & dev & test \\
        \midrule
        .216 & .211 & .216 & .211 \\[\bigtablesep]
        .651 & .620 & .495 & .523 \\[\bigtablesep]
        .717 & .673 & .680 & .665 \\[\bigtablesep]
        .738 & .709 & \textbf{.759} & .715 \\[\bigtablesep]
        \textbf{.744} & \textbf{.711} & .731 & \textbf{.725} \\
        \midrule
        .188 & .183 & .188 & .183 \\[\bigtablesep]
        .386 & .402 & .428 & .418 \\[\bigtablesep]
        .550 & .556 & .473 & .510 \\[\bigtablesep]
        .606 & .598 & \textbf{.628} & \textbf{.602} \\[\bigtablesep]
        \textbf{.627} & \textbf{.610} & .569 & .573 \\
        \midrule
        \multicolumn{4}{c}{.799} \\
    \end{tabular} \\
    \hhline{|-||-||-||-|}
\end{tabular}
}
\caption{Token $F_1$, segment $F_1$ and sentence $F_1$ on AURC-8 dev and test. Bold: best performance per column and split (in-domain, cross-domain).}
\tablabel{freq_base}
\end{table*}

\section{Results}
\tabref{freq_base} shows AURC results for our five methods (\secref{methods})
for in-domain (lines 4--8) and cross-domain (9--13).
The human upper bound (i.e., the performance on AURC by the three graduate 
employees) is given on line 14.
We report the mean of three runs with different random
seeds for non-deterministic methods \cite{tubiblio104568}.
Information about the hyperparameters, precision and recall, 
as well as separat evaluation for unit recognition (2 classes) and classification (3 classes) can be found in 
\secref{hyperparameters} and \secref{addinf} of the appendix.

\subsection{Model Comparison}

The BERT models always perform better than the majority baseline and FLAIR 
(6--8 vs.\ 4--5  and 11--13 vs.\ 9--10).
For
sentence-level \levelvariable\ and segment $F_1$,
the majority baseline is close 
or even better than FLAIR (lines 4 vs. 5 and 9 vs. 10).
The reason for the high performance of the majority baseline for segment $F_1$ (lines 4 \& 9) is  that about 40\% of segments are NON segments -- 
so a method that labels everything as NON will get a high score for 
segment identification.
BERT\textsubscript{BASE} mostly performs worse than BERT\textsubscript{LARGE}
(lines 6 vs. 7--8 and 11 vs. 12--13), but still at an
acceptable level.
Hence, BERT\textsubscript{BASE},
which requires only one GPU for  training,
is a good option
if computational resources are limited.

\subsection{Token- vs.\ Sentence-Level \Levelvariable}

BERT\textsubscript{LARGE}+CRF is consistently the best
model for token-level \levelvariable\ (bolded numbers in \tabref{freq_base}), whereas
BERT\textsubscript{LARGE} is the best model for
sentence-level \levelvariable\ (with one exception: for segment $F_1$ in-domain,
BERT\textsubscript{BASE} is better: line 6).
Thus,
for the token-level \levelvariable,
the CRF layer (lines 8 \& 13) always performs better than
imposing no sequence constraints (lines 7 \& 12) -- whereas the sequence constraint is not beneficial
for the sentence-level \levelvariable.

The difference between the two setups is noticeable for token and segment $F_1$, but smaller for
sentence $F_1$.
As one would expect, the token-level \levelvariable\
is not a  good match for sentence-level evaluation
(for both in- and cross-domain experiments).

\subsection{In-Domain vs.\ Cross-Domain}
The best in-domain BERT numbers are pretty high, with segment $F_1$ (line 8) even above human performance (line 14), while for token and sentence $F_1$ the gap to human performance is between 7 and 9 percentage points. 
Cross-domain  is clearly  more challenging
since models are evaluated on unseen topics. 
Here, $F_1$ scores drop to between 0.61 and 0.65 for the token-level models. 
Training on only 5 topics (see Section~\ref{sec:splits}) gives insufficient information for the transfer to unseen topics. 
These numbers are, however, in line with the current
state-of-the-art for sentence-level AM;
see \newcite{ReimersSBDSG19} who report an $F_1$ score of
0.63 for a similar architecture but slightly different
experimental setup with 7 training topics (cf. \newcite{Stab2018b}).
Given that the segment $F_1$ score (line 13) is only 6 percentage points below human agreement, we expect that in cases where exact argument segment boundaries are not of high importance, cross-domain transfer will still work fine.


\begin{figure*}[!th]
	\centering
	\includegraphics[width=0.99\textwidth]{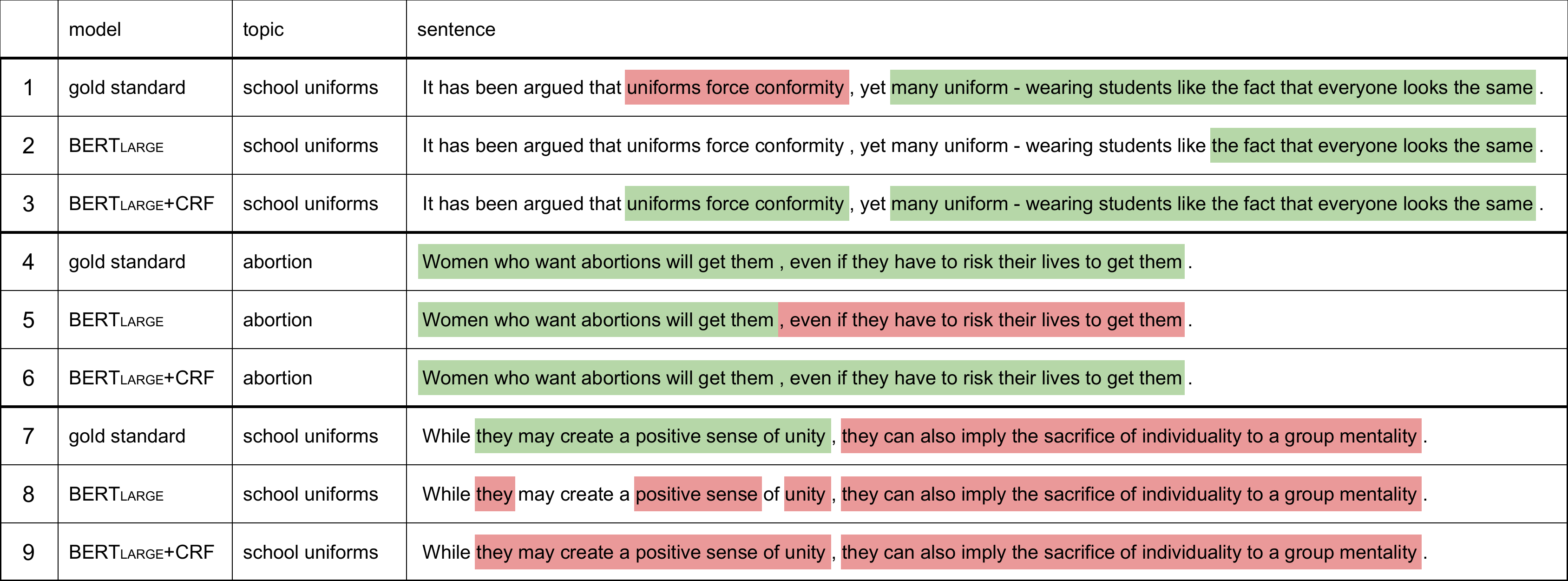}
    \caption{Gold standard segments and segments predicted by BERT\textsubscript{LARGE} and BERT\textsubscript{LARGE}+CRF for three example sentences. Green: PRO. Red: CON. }
	\label{fig:annotations}
\end{figure*}

\section{Analysis}
\seclabel{analysis}
We now discuss the main  errors
our models make (\secref{errors}) and 
investigate  robustness  against
incorrectly recognized sentence boundaries (\secref{robustness}).

\subsection{Recognition and Classification Errors}
\seclabel{errors}
The main  types of errors we observed were:
(i) the \emph{span} of an argumentative segment is not
correctly recognized and
(ii) the \emph{stance} is not correctly classified.

\textbf{Span.}
The first major type of error related to spans is that
the beginning and/or end of a segment is incorrectly recognized.
We analyzed the predictions by the best token-level model (line 8 \& 13 in \tabref{freq_base}) on  dev.
The average segment length (in tokens) is 17.5 for gold and 16.4 for predicted (in-domain) 
and 17.3 for gold and 15.1 for predicted (cross-domain); 
thus, predicted segments are on average shorter than gold segments.
However, this will often be benign in an application
since annotators
tended to include boundary words that are not strictly
necessary.

The second major type of error related to spans is 
that a segment is broken into several segments or several segments are merged into one segment.
Our models predict  7.6\% more segments in-domain and 13.1\% more segments
cross-domain than there are in the gold standard.
Shorter spans in the predictions
(which  result in more spans)
are mostly due to segment splits on function words like ``and'', ``that'' and ``yet''.
Figure \ref{fig:annotations} gives examples of span errors for the model without  CRF (lines 2, 5 \& 8).
The model with CRF (lines 3, 6 \& 8)
correctly recognizes beginning and end of the span more often, while the one without  CRF tends to split segments (line 5) and 
sometimes even creates one-word or two-word segments (line 8).
So the addition of a CRF layer improves segmentation
-- no overly short segments, fewer segments incorrectly
broken up --
and therefore results in the best overall performance in our experiments.

\textbf{Stance.}
The classification of stance is as important for argument
retrieval as span detection.  The model with CRF
sometimes assigns the same stance for all segments within a
sequence -- see the examples shown
in Figure \ref{fig:annotations} (lines 3 \& 9).  This may be due
to only a few mixed segments (both PRO and CON stance)
appearing in the training set ($43$ for in-domain and $48$
for cross-domain).
We plan to incorporate additional stance focused information -- such as sentiment
-- in future work to improve stance classification.
Another possible improvement could be achieved with multi-task training.
There are additional sentence-level datasets for stance detection \cite{mohammadsemeval} available.
In a multi-task setup, we can exploit these to improve stance detection for our task.

More detailed analysis of the subtask of recognition and classification 
(see \tabref{PRF123} in \secref{addinf} of the appendix)
showed that when combining PRO and CON labels into one ARG label for the best 
token-level system (BERT\textsubscript{LARGE}+CRF) yields a smaller drop in
performance between in- and cross-domain.
Consequently, our best token-level model manages cross-domain transfer very well for argument unit recognition, whereas argument (stance) classification remains challenging in a cross-domain setup.

\begin{table}\centering
\scriptsize
\scalebox{1.1}{
    \begin{tabular}{lrl|cc|cc}
        \toprule
         && \multicolumn{1}{r|}{}  & \multicolumn{2}{c|}{token-level} & \multicolumn{2}{c}{sentence-level} \\[\bigtablesep]
         && \multicolumn{1}{r|}{}  & \multicolumn{2}{c|}{\levelvariable} & \multicolumn{2}{c}{\levelvariable} \\
        \midrule
         && \multicolumn{1}{l|}{} & dev & test & dev & test \\
        \midrule
        \multirow{3}{*}{\rotatebox[origin=c]{90}{\scriptsize\begin{tabular}{c}in-\\domain\end{tabular}}}&
        3 & token $F_1$         & \textbf{.671} & \textbf{.642} & .507 & .510 \\[\bigtablesep]
 &       4 & segment $F_1$   & \textbf{.634} & \textbf{.592} & .254 & .235 \\[\bigtablesep]
  &      5 & sentence $F_1$  & \textbf{.514} & .504 & .510 & \textbf{.509} \\
        \midrule
        \multirow{3}{*}{\rotatebox[origin=c]{90}{\scriptsize\begin{tabular}{c}cross-\\domain\end{tabular}}}&
        6 & token $F_1$         & \textbf{.573} & \textbf{.562} & .442 & .452 \\[\bigtablesep]
&        7 & segment $F_1$    & \textbf{.554} & \textbf{.538} & .175 & .154 \\[\bigtablesep]
 &       8 & sentence $F_1$   & .433 & .449 & \textbf{.448} & \textbf{.473} \\
        \bottomrule
    \end{tabular}
}
\caption{\label{tab:sentence_boundaries}Token-level vs.\ sentence-level \levelvariable\
for input sequences that cross
        sentence boundaries (i.e., no sentence
        segmentation). Bold: best performance per
line and dev/test}
\end{table}

\subsection{Sentence Boundaries}
\seclabel{robustness}
One of the main benefits of a token-level argument mining
approach is that it is robust against errors in sentence
boundary detection. We now analyze how our
token-level models perform when they are not given
any sentence boundary information. We perform this
experiment with our best models
(BERT\textsubscript{LARGE}+CRF for token-level \levelvariable\ and 
BERT\textsubscript{LARGE} for sentence-level \levelvariable).

We create a new dataset by 
concatenating, for each topic, all sentences of that topic.
The resulting document is then processed by moving the model
over it 
using a sliding window of size 45  and a stepsize of 1.
In this regime,
most input sequences do not correspond to sentences, but
instead contain pieces of several sentences.
\tabref{sentence_boundaries} shows results.

We see that 
the token-level \levelvariable\ performs much better on input without
sentence boundaries
than the sentence-level \levelvariable\
-- both for token $F_1$ and for segment
$F_1$, both in-domain and cross-domain (lines 3-4, 6-7). 
While the performance is
clearly lower than
in \tabref{freq_base}, it
declines gracefully for the token-level \levelvariable\
whereas the drop in
performance for the sentence-level \levelvariable\ is large, especially
for segment $F_1$; e.g., on the test set: .154  (line
7, \tabref{sentence_boundaries}) vs.\ .473 (line
12, \tabref{freq_base}).
This clearly indicates the robustness of our approach
against errors in sentence boundary detection.


\section{Conclusion}
We define the task of fine-grained argument unit recognition and classification (AURC), 
and release the benchmark AURC-8 for this task.  
Our work is the first to show that information-seeking AM can be applied on very heterogeneous data (a web crawl in our case) on the level of tokens.
We demonstrated that AURC-8 has good quality in terms of annotator agreement: the
required annotations can be crowdsourced using specific data
selection and filtering methods combined with a slot filling
approach.  
We test and adapt several sequence tagging methods for AURC, achieving very good performance on known domains.
As in previous work on sentence-level argument retrieval, the methods suffer a drop in performance on unknown domains.  
Performance on unknown domains is, however,
quite good when segment boundaries do not need to be determined exactly.

We have also shown that our approach, as opposed to sentence-level argument retrieval, does not depend on correct sentence boundaries.
By making the analysis
more fine-grained, we increase the recall of arguments and
make their representation more accurate for the user as well
as for downstream tasks.
If required by downstream applications, the template defined in \secref{crowdannotations} can be used to reconstruct complete arguments for most of AURC-8.
For the example in Figure \ref{fig:annotations} line 7, we would get the following to statements:
``School uniforms should be supported because they may create a sense of positive unity'' and ``School uniforms should be opposed because they can also imply the sacrifice of individuality to a group mentally.''

This work makes possible a range of potential follow-up investigations. 
Since our approach does not depend on correct sentence boundaries, it is well-suited for noisy data from social media or (continuous) speech data, e.g., political debates.
Furthermore, the benefit of our fine-grained argument annotations should be explored in downstream applications like claim validation and fact checking, which rely on decomposed evidence.
Finally, our retrieval and annotation pipeline can be used to annotate argument units for additional topics, thus enabling improved domain transfer.


\section*{Acknowledgments}
We gratefully acknowledge support by Deutsche Forschungsgemeinschaft (DFG) 
(SPP-1999 Robust Argumentation Machines (RATIO), SCHU2246/13), 
as well as by the German Federal Ministry of Education and Research (BMBF) 
under the promotional reference 03VP02540 (ArgumenText).


\appendix

\section{Hyperparameters}
\seclabel{hyperparameters}

This section lists the hyperparameters used for the experimental systems 
described in the main part of the paper.

For FLAIR in the token-level model, we used a learning rate of 1e-1 with gradual 
decreasing, hiddensize=256 and for the sentence-level model the same setting for 
the learning rate, but with hiddensize=512.

For BERT\textsubscript{LARGE} and BERT\textsubscript{LARGE}+CRF, 
we used the large cased pretrained model with whole word masking 
and in the token-level setup a learning rate of 1e-5 for in- and cross-domain.
For the CRF we used the [CLS] token as the START and [SEP] as the END token,
so we considered only the sequence input (without the topic) for this setup.
For the evaluation we considered only the predicted tag sequence between the
START and END token.

We kept the learning rate at 4e-5 for the sentence-level 
BERT\textsubscript{LARGE} model and at 1e-5 for the 
BERT\textsubscript{LARGE}+CRF and used the AdamW optimizer.
The max. length of the tokenized BERT input was set to 64 tokens and we always
had a dropout rate of $0.1$.

All experiments were run three times with different seeds, 
a trainings batch size of 32 and for a max. of 100 epochs,
with earlier stopping if the performance/loss did not improve/decreased 
significantly (after ten epochs).

\section{Additional Information}
\seclabel{addinf}
\tabref{PRF123} provides detailed information about token-level precision, recall and 
macro $F_1$ scores and the separate evaluation on three classes (PRO vs. CON. vs. NON) 
and two classes (ARG vs. NON), where PRO and CON labels were converted into a single ARG label.

\begin{table}\centering
\scriptsize
\scalebox{1.1}{
    \begin{tabular}{lrrl|ccc}
        \toprule
        &   & domain       & set  & precision & recall & $F_1$ \\
        \midrule
        \multirow{4}{*}{\rotatebox[origin=c]{90}{\scriptsize\begin{tabular}{c} classific. \\ (3 classes)\end{tabular}}}
        & 2 & in-domain    & dev  & .742 & .745 & .743 \\[\bigtablesep]
        & 3 & cross-domain & dev  & .632 & .638 & .615 \\[\bigtablesep]
        & 4 & in-domain    & test & .717 & .681 & .696 \\[\bigtablesep]
        & 5 & cross-domain & test & .637 & .609 & .620 \\
        \midrule
        \multirow{4}{*}{\rotatebox[origin=c]{90}{\scriptsize\begin{tabular}{c}recogn. \\ (2 classes)\end{tabular}}}
        & 6 & in-domain    & dev  & .812 & .814 & .813 \\[\bigtablesep]
        & 7 & cross-domain & dev  & .805 & .792 & .797 \\[\bigtablesep]
        & 8 & in-domain    & test & .793 & .776 & .782 \\[\bigtablesep]
        & 9 & cross-domain & test & .782 & .767 & .770 \\
        \bottomrule
    \end{tabular}
}
\caption{\label{tab:PRF123}Token-level precision, recall and macro $F_1$ results 
    for argument classification (PRO, CON, NON) and recognition (ARG, NON) for the prediction of the best token-level model.}
\end{table}


\bibliography{AURC-AAAI20}

\begin{thebibliography}{}

\bibitem[\protect\citeauthoryear{Aharoni \bgroup et al\mbox.\egroup
  }{2014}]{aharoni-etal-2014-benchmark}
Aharoni, E.; Polnarov, A.; Lavee, T.; Hershcovich, D.; Levy, R.; Rinott, R.;
  Gutfreund, D.; and Slonim, N.
\newblock 2014.
\newblock A benchmark dataset for automatic detection of claims and evidence in
  the context of controversial topics.
\newblock In {\em Argumentation Mining Workshop},  64--68.

\bibitem[\protect\citeauthoryear{Ajjour \bgroup et al\mbox.\egroup
  }{2017}]{Ajjour2017}
Ajjour, Y.; Chen, W.-F.; Kiesel, J.; Wachsmuth, H.; and Stein, B.
\newblock 2017.
\newblock Unit segmentation of argumentative texts.
\newblock In {\em Argument Mining Workshop},  118--128.

\bibitem[\protect\citeauthoryear{Akbik, Blythe, and
  Vollgraf}{2018}]{akbik2018contextual}
Akbik, A.; Blythe, D.; and Vollgraf, R.
\newblock 2018.
\newblock Contextual string embeddings for sequence labeling.
\newblock In {\em COLING'18},  1638--1649.

\bibitem[\protect\citeauthoryear{Cabrio and Villata}{2018}]{ijcai2018-766}
Cabrio, E., and Villata, S.
\newblock 2018.
\newblock Five years of argument mining: a data-driven analysis.
\newblock In {\em IJCAI'18},  5427--5433.

\bibitem[\protect\citeauthoryear{Devlin \bgroup et al\mbox.\egroup
  }{2019}]{devlin2018bert}
Devlin, J.; Chang, M.-W.; Lee, K.; and Toutanova, K.
\newblock 2019.
\newblock {BERT}: Pre-training of deep bidirectional transformers for language
  understanding.
\newblock In {\em NAACL'19},  4171--4186.

\bibitem[\protect\citeauthoryear{Eckle-Kohler, Kluge, and
  Gurevych}{2015}]{EckleKohler2015}
Eckle-Kohler, J.; Kluge, R.; and Gurevych, I.
\newblock 2015.
\newblock On the role of discourse markers for discriminating claims and
  premises in argumentative discourse.
\newblock In {\em EMNLP'15},  2236--2242.

\bibitem[\protect\citeauthoryear{Habernal and Gurevych}{2017}]{habernal2017}
Habernal, I., and Gurevych, I.
\newblock 2017.
\newblock Argumentation mining in user-generated web discourse.
\newblock {\em Computational Linguistics} 43(1):125--179.

\bibitem[\protect\citeauthoryear{Hochreiter and
  Schmidhuber}{1997}]{hochreiter1997long}
Hochreiter, S., and Schmidhuber, J.
\newblock 1997.
\newblock Long short-term memory.
\newblock {\em Neural computation} 9(8):1735--1780.

\bibitem[\protect\citeauthoryear{Hovy \bgroup et al\mbox.\egroup
  }{2013}]{hovy2013learning}
Hovy, D.; Berg-Kirkpatrick, T.; Vaswani, A.; and Hovy, E.
\newblock 2013.
\newblock Learning whom to trust with mace.
\newblock In {\em NAACL'13},  1120--1130.

\bibitem[\protect\citeauthoryear{Hua and Wang}{2017}]{Hua2017}
Hua, X., and Wang, L.
\newblock 2017.
\newblock Understanding and detecting supporting arguments of diverse types.
\newblock In {\em ACL'17},  203--208.

\bibitem[\protect\citeauthoryear{Kim and Ghahramani}{2012}]{Kim2012IBCC}
Kim, H.-C., and Ghahramani, Z.
\newblock 2012.
\newblock Bayesian classifier combination.
\newblock In {\em AISTATS'12},  619--627.

\bibitem[\protect\citeauthoryear{Krippendorff \bgroup et al\mbox.\egroup
  }{2016}]{Krippendorff2016a}
Krippendorff, K.; Mathet, Y.; Bouvry, S.; and Widl\"{o}cher, A.
\newblock 2016.
\newblock On the reliability of unitizing textual continua: Further
  developments.
\newblock {\em Quality \& Quantity} 50(6):2347--2364.

\bibitem[\protect\citeauthoryear{Kuribayashi \bgroup et al\mbox.\egroup
  }{2019}]{kuribayashi-etal-2019-empirical}
Kuribayashi, T.; Ouchi, H.; Inoue, N.; Reisert, P.; Miyoshi, T.; Suzuki, J.;
  and Inui, K.
\newblock 2019.
\newblock An empirical study of span representations in argumentation structure
  parsing.
\newblock In {\em ACL'19},  4691--4698.

\bibitem[\protect\citeauthoryear{Lafferty, McCallum, and
  Pereira}{2001}]{lafferty2001conditional}
Lafferty, J.~D.; McCallum, A.; and Pereira, F. C.~N.
\newblock 2001.
\newblock Conditional random fields: Probabilistic models for segmenting and
  labeling sequence data.
\newblock In {\em ICML'01},  282--289.

\bibitem[\protect\citeauthoryear{Lample \bgroup et al\mbox.\egroup
  }{2016}]{lample2016neural}
Lample, G.; Ballesteros, M.; Subramanian, S.; Kawakami, K.; and Dyer, C.
\newblock 2016.
\newblock Neural architectures for named entity recognition.
\newblock In {\em NAACL'16},  260--270.

\bibitem[\protect\citeauthoryear{Landis and Koch}{1977}]{Landis1977}
Landis, J.~R., and Koch, G.~G.
\newblock 1977.
\newblock The measurement of observer agreement for categorical data.
\newblock {\em Biometrics} 33(1):159--174.

\bibitem[\protect\citeauthoryear{Levy \bgroup et al\mbox.\egroup
  }{2014}]{Levy2014}
Levy, R.; Bilu, Y.; Hershcovich, D.; Aharoni, E.; and Slonim, N.
\newblock 2014.
\newblock Context dependent claim detection.
\newblock In {\em COLING'14},  1489--1500.

\bibitem[\protect\citeauthoryear{Li \bgroup et al\mbox.\egroup }{2017}]{Li2017}
Li, M.; Geng, S.; Gao, Y.; Peng, S.; Liu, H.; and Wang, H.
\newblock 2017.
\newblock Crowdsourcing argumentation structures in {Chinese} hotel reviews.
\newblock In {\em IEEE Int. Conference on Systems, Man, and Cybernetics},
  87--92.

\bibitem[\protect\citeauthoryear{Lippi and
  Torroni}{2016}]{Lippi:2016:AMS:3016100.3016319}
Lippi, M., and Torroni, P.
\newblock 2016.
\newblock Argument mining from speech: Detecting claims in political debates.
\newblock In {\em AAAI'16},  2979--2985.

\bibitem[\protect\citeauthoryear{Miller, Sukhareva, and
  Gurevych}{2019}]{miller2019}
Miller, T.; Sukhareva, M.; and Gurevych, I.
\newblock 2019.
\newblock A streamlined method for sourcing discourse-level argumentation
  annotations from the crowd.
\newblock In {\em NAACL'19},  1790--1796.

\bibitem[\protect\citeauthoryear{Mohammad \bgroup et al\mbox.\egroup
  }{2016}]{mohammadsemeval}
Mohammad, S.; Kiritchenko, S.; Sobhani, P.; Zhu, X.; and Cherry, C.
\newblock 2016.
\newblock {S}em{E}val-2016 task 6: Detecting stance in tweets.
\newblock In {\em {S}em{E}val'16},  31--41.

\bibitem[\protect\citeauthoryear{Nguyen and
  Litman}{2018}]{DBLP:conf/aaai/NguyenL18}
Nguyen, H.~V., and Litman, D.~J.
\newblock 2018.
\newblock Argument mining for improving the automated scoring of persuasive
  essays.
\newblock In {\em AAAI'18},  5892--5899.

\bibitem[\protect\citeauthoryear{Palau and Moens}{2009}]{Palau2009}
Palau, R.~M., and Moens, M.-F.
\newblock 2009.
\newblock Argumentation mining: The detection, classification and structure of
  arguments in text.
\newblock In {\em ICAIL'09},  98--107.

\bibitem[\protect\citeauthoryear{Peldszus and Stede}{2013}]{Peldszus2013}
Peldszus, A., and Stede, M.
\newblock 2013.
\newblock From argument diagrams to argumentation mining in texts: A survey.
\newblock {\em Int. J. Cogn. Inform. Nat. Intell.} 7(1):1--31.

\bibitem[\protect\citeauthoryear{Peldszus and
  Stede}{2015}]{peldszus-stede-2015-joint}
Peldszus, A., and Stede, M.
\newblock 2015.
\newblock Joint prediction in {MST}-style discourse parsing for argumentation
  mining.
\newblock In {\em Proceedings of the 2015 Conference on Empirical Methods in
  Natural Language Processing},  938--948.
\newblock Lisbon, Portugal: Association for Computational Linguistics.

\bibitem[\protect\citeauthoryear{Pennington, Socher, and
  Manning}{2014}]{pennington2014glove}
Pennington, J.; Socher, R.; and Manning, C.
\newblock 2014.
\newblock Glove: Global vectors for word representation.
\newblock In {\em EMNLP'14},  1532--1543.

\bibitem[\protect\citeauthoryear{Reimers and Gurevych}{2017}]{tubiblio104568}
Reimers, N., and Gurevych, I.
\newblock 2017.
\newblock Reporting score distributions makes a difference: Performance study
  of lstm-networks for sequence tagging.
\newblock In {\em EMNLP'17},  338--348.

\bibitem[\protect\citeauthoryear{Reimers \bgroup et al\mbox.\egroup
  }{2019}]{ReimersSBDSG19}
Reimers, N.; Schiller, B.; Beck, T.; Daxenberger, J.; Stab, C.; and Gurevych,
  I.
\newblock 2019.
\newblock Classification and clustering of arguments with contextualized word
  embeddings.
\newblock In {\em ACL'19},  567--578.

\bibitem[\protect\citeauthoryear{Reisert \bgroup et al\mbox.\egroup
  }{2018}]{Reisert2018}
Reisert, P.; Inoue, N.; Kuribayashi, T.; and Inui, K.
\newblock 2018.
\newblock Feasible annotation scheme for capturing policy argument reasoning
  using argument templates.
\newblock In {\em Argument Mining Workshop},  79--89.

\bibitem[\protect\citeauthoryear{Shnarch \bgroup et al\mbox.\egroup
  }{2018}]{Shnarch2018}
Shnarch, E.; Alzate, C.; Dankin, L.; Gleize, M.; Hou, Y.; Choshen, L.;
  Aharonov, R.; and Slonim, N.
\newblock 2018.
\newblock Will it blend? blending weak and strong labeled data in a neural
  network for argumentation mining.
\newblock In {\em ACL'18},  599--605.

\bibitem[\protect\citeauthoryear{Simpson and Gurevych}{2019}]{Simpson2018}
Simpson, E.~D., and Gurevych, I.
\newblock 2019.
\newblock A {B}ayesian approach for sequence tagging with crowds.
\newblock In {\em EMNLP-IJCNLP'19},  1093--1104.

\bibitem[\protect\citeauthoryear{Stab and Gurevych}{2014}]{Stab2014a}
Stab, C., and Gurevych, I.
\newblock 2014.
\newblock Annotating argument components and relations in persuasive essays.
\newblock In {\em COLING'14},  1501--1510.

\bibitem[\protect\citeauthoryear{Stab \bgroup et al\mbox.\egroup
  }{2018a}]{stab2018argumentext}
Stab, C.; Daxenberger, J.; Stahlhut, C.; Miller, T.; Schiller, B.; Tauchmann,
  C.; Eger, S.; and Gurevych, I.
\newblock 2018a.
\newblock Argumentext: Searching for arguments in heterogeneous sources.
\newblock In {\em NAACL'18},  21--25.

\bibitem[\protect\citeauthoryear{Stab \bgroup et al\mbox.\egroup
  }{2018b}]{Stab2018b}
Stab, C.; Miller, T.; Schiller, B.; Rai, P.; and Gurevych, I.
\newblock 2018b.
\newblock Cross-topic argument mining from heterogeneous sources.
\newblock In {\em EMNLP'18},  3664--3674.

\bibitem[\protect\citeauthoryear{Wachsmuth \bgroup et al\mbox.\egroup
  }{2017}]{Wachsmuth2017}
Wachsmuth, H.; Potthast, M.; Al~Khatib, K.; Ajjour, Y.; Puschmann, J.; Qu, J.;
  Dorsch, J.; Morari, V.; Bevendorff, J.; and Stein, B.
\newblock 2017.
\newblock Building an argument search engine for the web.
\newblock In {\em Argument Mining Workshop},  49--59.

\bibitem[\protect\citeauthoryear{Walton and Gordon}{2017}]{DouglasWalton2017}
Walton, D., and Gordon, T.~F.
\newblock 2017.
\newblock {Argument Invention with the Carneades Argumentation System}.
\newblock {\em SCRIPTed} 14(2):168--207.

\bibitem[\protect\citeauthoryear{Wyner \bgroup et al\mbox.\egroup
  }{2010}]{Wyner2010}
Wyner, A.; Mochales-Palau, R.; Moens, M.-F.; and Milward, D.
\newblock 2010.
\newblock Approaches to text mining arguments from legal cases.
\newblock In {\em Semantic Processing of Legal Texts: Where the Language of Law
  Meets the Law of Language}. Springer Berlin Heidelberg.
\newblock  60--79.

\end{thebibliography}
\bibliographystyle{aaai}


\end{document}